\documentclass[letterpaper, 10 pt, conference]{ieeeconf}  
\IEEEoverridecommandlockouts                                                                                
\usepackage{graphics}
\usepackage{graphicx}
\usepackage{amssymb} 
\usepackage{lipsum}
\usepackage{color}
\usepackage{amsmath}
\usepackage{siunitx}
\usepackage{multirow}
\usepackage{multicol}
\usepackage{cite}
\usepackage[belowskip=-1pt,aboveskip=5pt]{caption}
\usepackage{adjustbox}
\usepackage{float}
\usepackage{tikz}
\newcommand\copyrighttext{
	\footnotesize \textcopyright 2022 IEEE. Personal use of this material is permitted.  Permission from IEEE must be obtained for all other uses, in any current or future media, including reprinting/republishing this material for advertising or promotional purposes, creating new collective works, for resale or redistribution to servers or lists, or reuse of any copyrighted component of this work in other works.
	DOI: 10.1109/ITSC55140.2022.9922148}
\newcommand\copyrightnotice{
	\begin{tikzpicture}[remember picture,overlay]
		\node[anchor=south,yshift=10pt] at (current page.south) {\fbox{\parbox{\dimexpr\textwidth-\fboxsep-\fboxrule\relax}{\copyrighttext}}};
	\end{tikzpicture}
}

\title{\LARGE \bf
Short-term Inland Vessel Trajectory Prediction \\with Encoder-Decoder Models \\}

\author{Kathrin Donandt$^{1}$, Karim Böttger$^{2}$, Dirk Söffker$^{3}$
\thanks{$^{1}$K. Donandt is with the Institute of Ship Technology, Ocean Engineering and Transport Systems (ISMT), University of Duisburg-Essen, Duisburg, Germany. 
        {\tt\small kathrin.donandt@uni-due.de}}%
\thanks{$^{2}$K. Böttger is with the German Federal Waterway Engineering and Research Institute.
        {\tt\small karim.boettger@baw.de}}%
\thanks{$^{3}$D. Söffker is with the Chair of Dynamics and Control (SRS), University of Duisburg-Essen, Duisburg, Germany.
        {\tt\small soeffker@uni-due.de}}
\thanks{*This publication originates from a joint research project between the
German Federal Waterways Engineering and Research Institute and the ISMT,
and was written in collaboration with the SRS.}
}

\begin{document}

\maketitle
\copyrightnotice
\thispagestyle{empty}
\pagestyle{empty}

\begin{abstract}
Accurate vessel trajectory prediction is necessary for save and efficient navigation. Deep learning-based prediction models, esp. encoder-decoders, are rarely applied to inland navigation specifically. Approaches from the maritime domain cannot directly be transferred to river navigation due to specific driving behavior influencing factors. Different encoder-decoder architectures, including a transformer encoder-decoder, are compared herein for predicting the next positions of inland vessels, given not only spatio-temporal information from AIS, but also river specific features. The results show that the reformulation of the regression task as classification problem and the inclusion of river specific features yield the lowest displacement errors. The standard LSTM encoder-decoder outperforms the transformer encoder-decoder for the data considered, but is computationally more expensive. In this study for the first time a transformer-based encoder-decoder model is applied to the problem of predicting the ship trajectory. Here, a feature vector using the river-specific context of navigation input parameters is established. Future studies can built on the proposed models, investigate the improvement of the  computationally more efficient transformer, e.g. through further hyper-parameter optimization, and use additional river-specific information in the context representation to further increase prediction accuracy.

\end{abstract}

\begin{keywords}
vessel trajectory prediction, encoder-decoder model, transformer, inland navigation
\end{keywords}

\section{INTRODUCTION}
Predicting the driving behavior of vessels accurately is crucial to avoid collisions and find an efficient route when navigating in confined spaces. 
Automatic prediction can support navigators whereas autonomous vessels are completely dependent on it to determine the path to follow.
Ship models, which can predict a vessel's trajectory with high precision by  considering the individual ship's driving dynamics in a given waterway condition, are expensive to be obtained and are mostly used in simulators. In real world applications that include trajectory prediction, simple physical models, such as constant velocity models (CVM), still prevail \cite{Murray.2021b, Capobianco.2021}. Kalman filters have been suggested to overcome the linearity constraint of CVMs but have limited prediction horizons \cite{Murray.2021b}.
With the rising AIS data availability in the last years, data-driven models, including deep neural networks, have gained popularity. They have been studied as an alternative or in addition to model-based approaches to deal with complex navigation situations and longer prediction horizons. 
Deep learning (DL)-based trajectory prediction has mainly been applied to ocean-going vessels using spatio-temporal vessel data alone. This is problematic in the inland shipping domain, where waterway characteristics such as river flow and river shape restrict the vessels' navigation. 
In this study, such contextual information is taken into consideration by adapting the common initial decoder input of encoder-decoder models. 
Encoder-decoder models have in general shown superior performance for sequence-to-sequence tasks compared to classical recurrent neural networks (RNNs). The advantage of these models over RNNs which require a recursive strategy to predict several time steps into the future has already been confirmed for the inland domain as well (see \cite{You.2020}). Therefore, encoder-decoder models are used herein to predict a sequence of next positions of a vessel's trajectory given a sequence of past positions. 
Transformer encoder-decoder models, that have not yet been applied to trajectory prediction in navigation\footnote{The study of \cite{Nguyen.2021TrAISformerAGT}, see Sec. \ref{sec:relatedwork}. for details, has not yet been published.}, are investigated. 

A brief overview of related works in the field of DL-based trajectory prediction for vessels specifically, and transformer-based approaches for trajectory prediction in general, is given in the following section. In Section \ref{sec:background}, encoder-decoders are shortly explained before the inland-specific, context-sensitive classification transformer is introduced in Section \ref{sec:approach}. Section \ref{sec:results} is dedicated to the experiment details, evaluation methods and results, before a conclusion and outlook is given in Section \ref{sec:conclusion}.

\section{RELATED WORK}\label{sec:relatedwork}
\subsection{Deep learning-based vessel trajectory prediction}
Deep learning has been successfully applied for trajectory prediction of humans, vehicles and maritime vessels \cite{Rudenko.2020, Mozaffari.2020, Tu.2018, Azimi.2020, Xiao.2020,Jin.2021}. Inland navigation-specific DL approaches, are, however, rare. Encoder-decoder architectures have gained popularity in the last years in the navigation domain \cite{Nguyen.2018,You.2020,Forti.2020,Sekhon.2020,DijtPimandMettesPascal.2020,Capobianco.2021}. For inland navigation specifically, only two studies \cite{You.2020,DijtPimandMettesPascal.2020} exist in which this neural network architecture is suggested. 
In \cite{You.2020}, different RNN encoder-decoder models without attention process spatio-temporal data of an inland vessel of the last 10 min to predict its positions for the upcoming 5 min. The proposed model outperforms the RNN baseline which predicts the output sequence by repeatedly feeding the last output back to the model as new input sequence element.

Waterway specific information in addition to spatio-temporal vessel data has rarely been considered in the literature on DL-based vessel trajectory prediction. In \cite{DijtPimandMettesPascal.2020}, radar images are mapped to electronic navigational charts (ENC) by a convolutional neural network in order to generate a feature vector representing the navigation context. This feature vector is then fed to an LSTM encoder-decoder for inland vessel trajectory prediction. 

No homogeneous understanding of ``short-term'' and ``long-term'' prediction in inland navigation exist in the literature covering DL-based approaches. For anticipation of future ship locations in the context of autonomous navigation, \cite{DijtPimandMettesPascal.2020} use input and output sequences comprising 15 s and 150 s, respectively, and a time interval size between consecutive time steps of 3 s. The authors speak of ``long-term'' prediction probably due to the high number of prediction steps with the given time step size. 
In \cite{You.2020}, short-term prediction horizons are defined to range from 5 to 15 min in the context of collision risk alert, the time interval size is set to 30 s and an input sequence of 20 data points (10 min) is fed to the predictor. Finally, a trajectory segment of up to 10 min is reconstructed given an input sequence of 4 min before or after this segment, using a time interval size of 1 min \cite{Yuan.2020}.

In the present work, the observation and prediction windows are defined similar to \cite{You.2020} and \cite{Yuan.2020}. Different input and output sequence lengths combinations are tested. During training, either 5 or 10 min are fed to the prediction model to obtain the next 15 or 30 min. Thus, the effect of varying the input length as well as the output length can be analyzed. The unusually long prediction horizon of 30 min is only used for model training, and the errors in a prediction window of up to 15 min are compared to the errors of models trained with smaller prediction windows.

\subsection{Transformer for trajectory prediction}
Transformer encoder-decoder models based on attention mechanisms \cite{Vaswani.2017} have shown superior performance when compared to the standard long short-term memory (LSTM) encoder-decoder models in the Natural Language Processing field, where it has first been applied. It has already been employed to trajectory prediction for humans and automotive vehicles \cite{Cai.2020, Chen.2021,Yu.2020}. For vessel trajectory prediction, a transformer that predicts a single time step ahead is presented in \cite{Nguyen.2021TrAISformerAGT}. A transformer encoder-decoder model mapping an input to an output time series has not been investigated yet.

\section{METHODOLOGICAL BACKGROUND}\label{sec:background}
\begin{figure}
\centering
{\includegraphics[width=0.48\textwidth,clip,height=0.27\textheight]{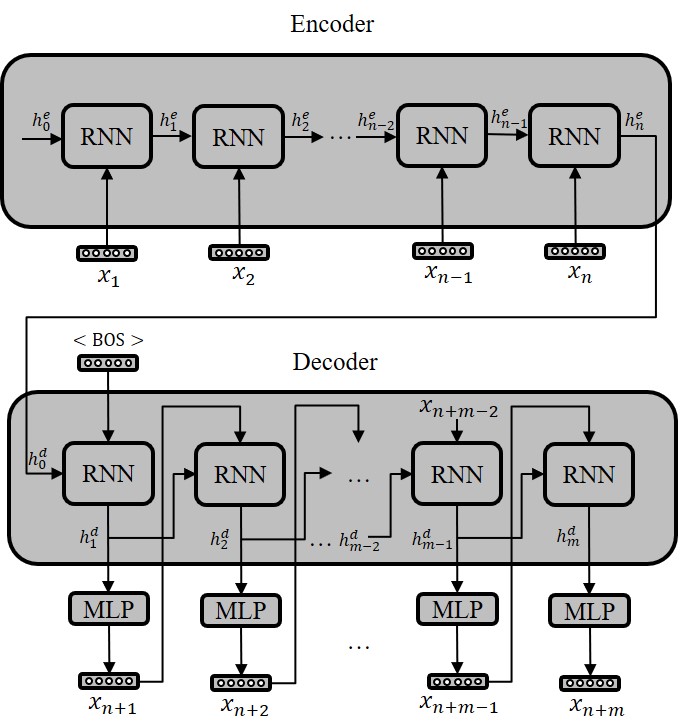}}
	\caption{Classical encoder-decoder architecture}
	\label{fig:EncDec}
\end{figure}

Encoder-decoder models have originally been developed for machine translation\cite{Sutskever.2014}, but are nowadays applied to other sequence-to-sequence modelling problems as well. 
The architecture of a classical encoder-decoder model is depicted in Fig. \ref{fig:EncDec}.
The information of the whole input sequence $x_1,...,x_n$ is encoded by an RNN into a so-called context-vector which is fed to the decoder as first hidden state $h_0^d$. In the classical model, the context vector simply consists of the last encoder hidden state such that $h_n^e = h_0^d$ . The decoder generates the output sequence $x_{n+1},...,x_{n+m}$ by another RNN, receiving as input at time step $t$ the previous decoder output. The first decoder input commonly consists in a place-holder (commonly referred to as $<$BOS$>$ (``beginning-of-sequence'') token). The hidden state obtained after processing one time step in the decoder-RNN is fed to the output layer (e.g. a multilayer perceptron (MLP)) to obtain the prediction. More advanced encoder-decoders use aggregation functions that process all hidden states of the encoder to generate the context vector. This is expected to improve the usage of information from the encoded sequence in the decoder. Attention-based aggregation functions are commonly used. They are able to consider the individual significance of each input when generating the output sequence. Transformers\cite{Vaswani.2017} are state of the art encoder-decoders substituting the encoder- and decoder-RNNs by stacked attention and feedforward layers. The sequentiality of the input and output sequences is taken into consideration by positional encodings. 

\section{CONTEXT-SENSITIVE CLASSIFICATION TRANSFORMER}\label{sec:approach}
\begin{figure}
\centering
{\includegraphics[width=0.5\textwidth,clip,keepaspectratio]{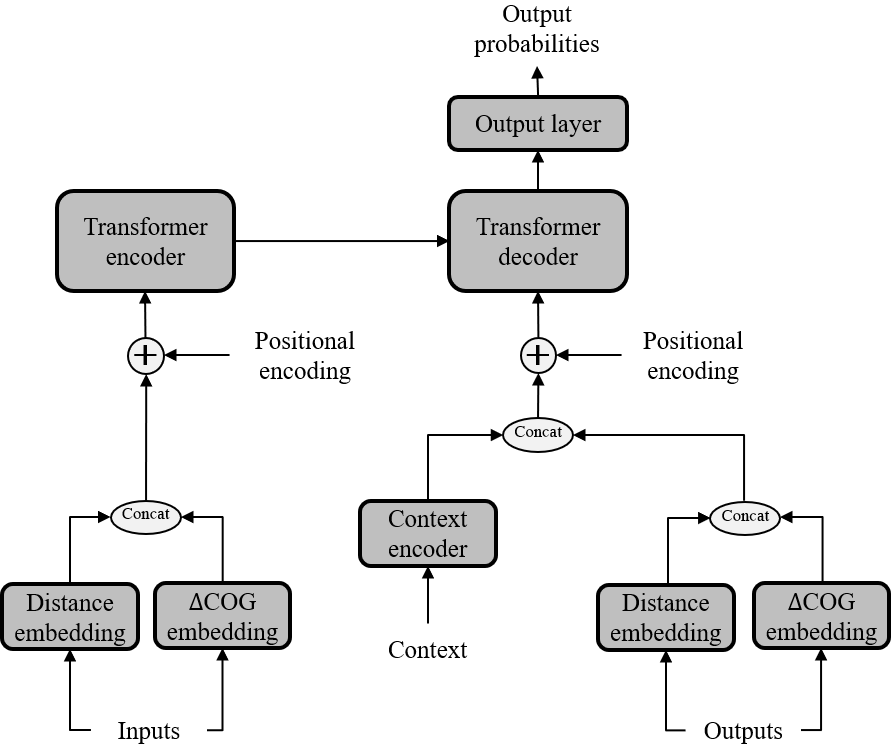}}
	\caption{Context-sensitive classification transformer}
	\label{fig:transformer}
\end{figure}

The overall architecture of the transformer encoder-decoder model for inland vessel trajectory prediction developed in this study is shown in Fig. \ref{fig:transformer}. It differs from the general transformer architecture only in the input and output sequence processing layers preceding the actual transformer-specific model parts (positional encoding, transformer encoder and decoder). For details on the inner structure of these transformer-specific components, the interested reader is referred to \cite{Vaswani.2017}. The proposed model receives a time series of $n \in \{5,10\}$  travelled distances and course over ground (COG) changes as input, and generates the next $m \in \{ 15,30\}$ distances and COG changes. The time interval size is 1 min, and three combinations of $n$ and $m$ are considered (5-15, 10-15 and 10-30). A sequence fed to the encoder-decoder model is defined as 
    \[
    x_1,...,x_{n+m} ,
    \] 
where
\[
x_t = (|pos_{t+1}-pos_t|, cogDi\!f\!f(t)) 
\]
\[
\forall t \in \{1,...,n+m\}
\]
and
    \[
    cogDi\!f\!f(t) 
\begin{cases}
    + w, & \text{clock-wise change}\\
    - w, & \text{counter-clock-wise change}
\end{cases}
\]
with
\[
 w = min\{|cog_{t+1}-cog_t|, 360-|cog_{t+1}-cog_t|\}.
\]

For simplicity, 180° is assumed to be the maximum change of the course over ground (COG) in two consecutive time steps. 
The distance and $\Delta$COG embedding layers encode $|pos_{t+1}-pos_t|$ and $cogDi\!f\!f(t)$, respectively, into high-dimensional feature vectors of these input and output components.

The original regression problem of predicting the next position is reformulated as classification problem, thus the output sequence is obtained from output probabilities of classes and not directly. 
Due to this reformulation, an adequate loss function for classification models needs to be employed. The overall loss of the proposed classification model is therefore defined as 

\[
\mathcal{L}_{total} = \frac{1}{2} CEL_{cogDi\!f\!f} + \frac{1}{2} CEL_{distance},
\]

where $CEL_{cogDi\!f\!f}$ and $CEL_{distance}$ calculate the cross entropy loss (CEL) of the predicted probability distributions for the COG changes and the distances, respectively.

The classification reformulation is inspired by the works of Nguyen et al. \cite{Nguyen.2018, Nguyen.2021TrAISformerAGT}. 
By obtaining a probability distribution as output, it is possible to sample several possible output sequences and determine the model's uncertainty. 
The encoder-decoder model could thus be part of hybrid model that falls back to an alternative prediction model in case of high uncertainty. 
In this study, a greedy decoding method is used for simplicity instead of sampling. It picks the most likely output class at each predicted time step to generate a single output sequence per input. 

To obtain the class labels, the continuous distance and COG change values are discretized with a resolution of 1 m or 10 m (depending on the prediction horizon) and 1°, respectively. The euclidean distance between the last position $pos_{n+m+1}$ of the original (interpolated) AIS data and the last position $pos'_{n+m+1}$, which is obtained by calculating the coordinates of the sequence using the initial position $pos_1$, and the $n+m$ discretized distance and COG change values, is in average 25 m after 20 min and 70 m after 40 min. This is still less than the length of most commercial inland vessels. The deviation of the trajectory obtained with discretized distances and COG changes from the trajectory of the AIS data can be seen in Fig. \ref{fig:predictionEx} (blue and green lines).

The $<$BOS$>$ token commonly used as first input to the decoder is replaced by the last encoder input in \cite{Capobianco.2021}, where a bi-directional LSTM encoder-decoder is proposed for vessel trajectory prediction. In the present study, a vector representation of river-specific information is taken instead. It is expected that the inclusion of such context information improves the prediction performance, as inland navigation is constrained by several and frequently changing waterway conditions. Whereas only river radii are considered in this study, the approach is not limited to this specific river features, but can be extended to also consider water levels, flow velocities, fairway widths etc.
For a specific sequence $x_1,...,x_{n+m}$, the context information vector is defined as
    \[
    (cv(hm(pos_1)+0.1\times 0),
    cv(hm(pos_1)+0.1\times 1),...,\]
    \[
    cv(hm(pos_1)+0.1\times N)), 
    \]    with
    \[
    N = (7.7 \times (m+n) \times 60) / 100,
    \]
where $cv$ is a function returning the river curvature value given a specific hectometer, and $hm$ is the hectometer function returning the hectometer for a given position. Thus, the context information is a sequence of $cv$ values for each 100 m starting from the hectometer of the sequence's first position $hm(pos_1)$ and ending at the hectometer that can be reached within $n+m$ min with a maximum speed of 15 knots (7.7 m/s).
For simplicity, it is assumed that the difference between this hectometer range and the real travelled distance is neglectable in the context of river-specific information representation. 
The context encoder consists of an embedding and feedforward layer.
The curvature function values are discretized and passed through the embedding layer. The resulting $N$ embeddings are concatenated and passed through the feedforward layer to obtain the final context vector representation used as first decoder input.

\section{EXPERIMENTS AND RESULTS}\label{sec:results}
\subsection{Data details and pre-processing}\label{Data}
The data set used in this study was provided by the German Federal Waterway Engineering and Research Institute (BAW) and comprises 29 months (05/2019 - 09/2021) of AIS data of a free-flowing section of the Danube river (hectometer range 2231.4 to 2321.44). A number of 7.7k trajectories with increasing hectometer direction are extracted (no turning allowed). As the vessel driving behaviors in free-flowing rivers differ significantly depending on the trajectory direction in relation to the river flow (up- or downhill), direction-specific models are developed in this study. 

The time steps are standardized through linear interpolation, resulting in constant velocity and course over ground between consecutive AIS data points. The error caused by this interpolation is assumed to be neglectable for time intervals of up to 120 s. No interpolation is applied between consecutive data points with a larger time interval. 
From the interpolated trajectories, between 250k and 280k sequences of 20, 25, and 40 data points are extracted using a sliding window. Trajectory parts for which no interpolation is possible are not considered. The obtained sequences are split into training (80\%), validation (10\%) and test set (10\%). Sequences containing time intervals of more than 2 min (i.e. trajectory parts that could not be interpolated) are not allowed. 

For $pos_t$, the AIS latitude and longitude values are converted to planar coordinates to enable euclidean distance calculations. The $cog_t$ values are calculated given the planar coordinates and not taken from the interpolated AIS data directly to avoid inconsistencies.

The radii for each hectometer necessary to generate the context vector (see Sec. \ref{sec:approach}) were also provided by BAW. The river curvature function $cv$ is obtained by inverting the radii values (to avoid infinity values) and interpolation. The hectometer function $hm$ uses the hectometers provided by BAW for each AIS data point.
\begin{figure}
\centering
{\includegraphics[width=0.5\textwidth,clip,height=0.25\textheight]{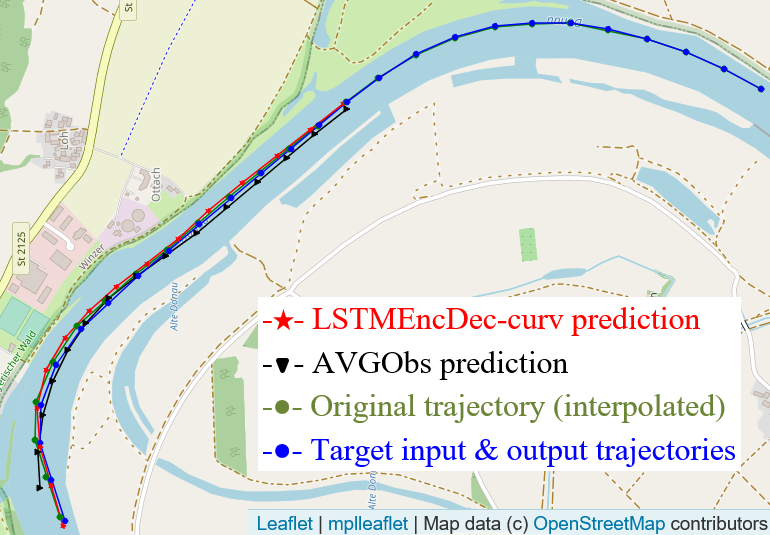}}
	\caption{Prediction example for $n = 10$ and $m = 15$.}
	\label{fig:predictionEx}
\end{figure}

\subsection{Model variants}
The classification transformer model described in Sec. \ref{sec:approach} is compared to a combination of a regression and a classification model. 
The loss of this combined model is defined as 
\begin{equation}
  \begin{aligned}
\mathcal{L}_{total} & = \alpha\left(\frac{1}{2}CEL_{cogDi\!f\!f} + \frac{1}{2} CEL_{distance}\right) + \\ 
& (1-\alpha)\left(\frac{1}{2}MSE_{cogDi\!f\!f} + \frac{1}{2} MSE_{distance}\right),
\nonumber
\end{aligned}
\end{equation}

where $CEL_{cogDi\!f\!f}$ and $CEL_{distance}$ are the cross entropy losses of the classification sub-model's predictions, $MSE_{cogDi\!f\!f}$ and $MSE_{distance}$ the mean square errors of the regression sub-model's predictions, and $\alpha$ is a weighting factor and set to 0.5 in this study. 
The motivation of adding a regression sub-module is to give the classification model a notion of the numerical relation between the class labels. Typical classification problems do not assume a numerical relation between class labels. 

To determine the performance gain when using a vector representation of the river curvature values as initial decoder input, the context-agnostic counterpart of the suggested model (i.e. using the default $<$BOS$>$ embedding as initial decoder input) is also tested.

\subsection{Comparison to LSTM encoder-decoder}
A LSTM encoder-decoder without any aggregation function is compared to the suggested transformer model. The input and output sequences as well as the context information are processed by the corresponding embedding layers and context encoder as in the transformer (see Fig. \ref{fig:transformer}) before being fed to the encoder and decoder. The loss function is identical to the one proposed for the transformer, and a context-agnostic LSTM encoder-decoder is tested as well.

\begin{figure}
\centering
{\includegraphics[width=0.47\textwidth,clip,height=0.22\textheight]{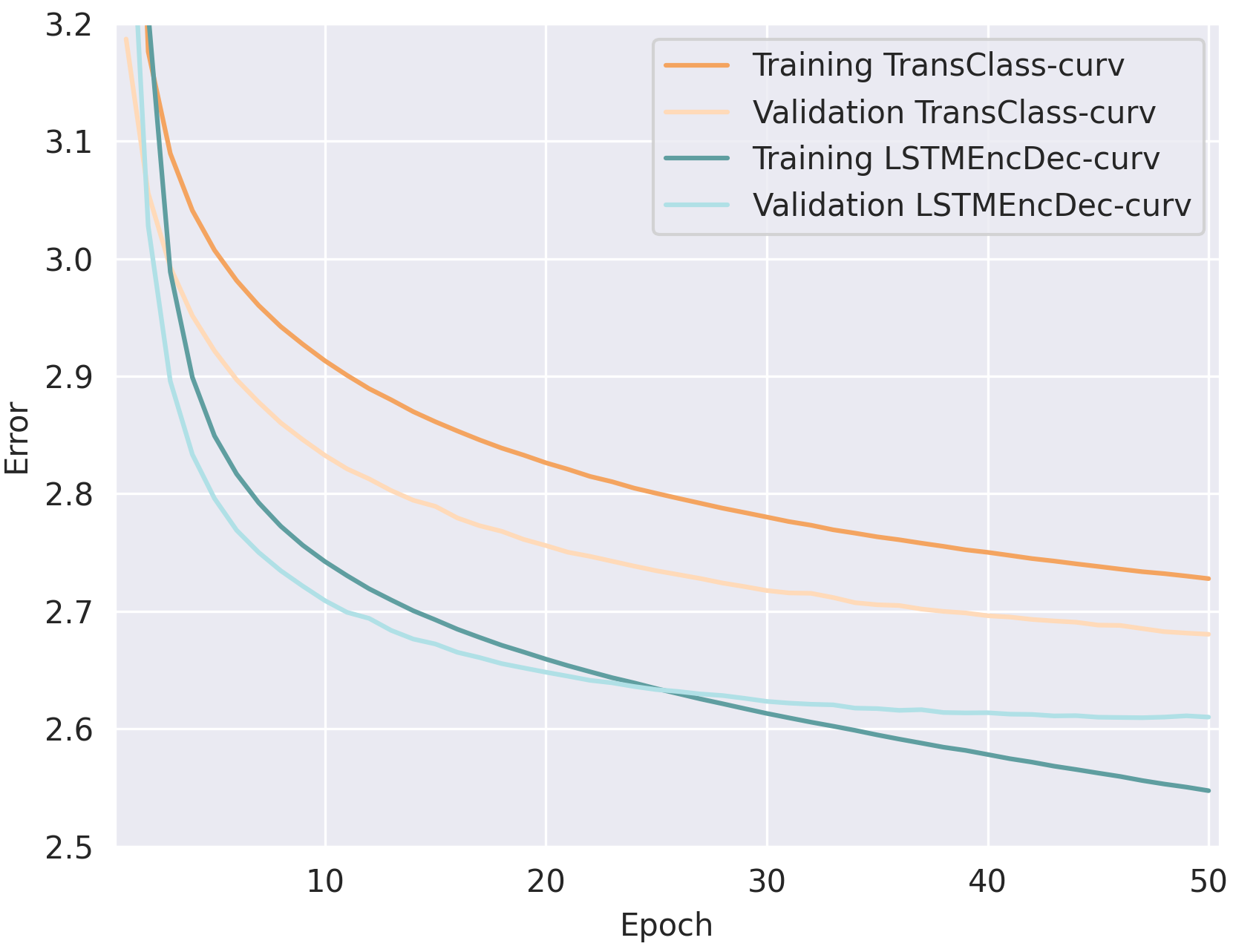}}
	\caption{Training and validation loss  curves.}\label{fig:ValidTrans}
\end{figure}
\subsection{Model training}
For both the context-sensitive LSTM and transformer encoder-decoder (LSTMEncDec-curv and TransClass-curv) approaches, the hyper-parameters have to be defined. The best performance is obtained with the following configurations: 
\begin{itemize}
    \item Dropout: 0.1
    \item Hidden layer size (i.e. the number of hidden state features of the LSTM and the size of the fully connected layer in the transformer): 512
    \item Number of encoder/decoder layers: 3/3
    \item Size of distance/$\Delta$COG embedding (cf. Fig. \ref{fig:transformer}): 512/512 (LSTMEncDec-curv) and 256/256 (TransClass-curv)
    \item Optimizer: Adam
\end{itemize}
 
\begin{figure}[]
{\includegraphics[width=0.48\textwidth,clip,keepaspectratio]{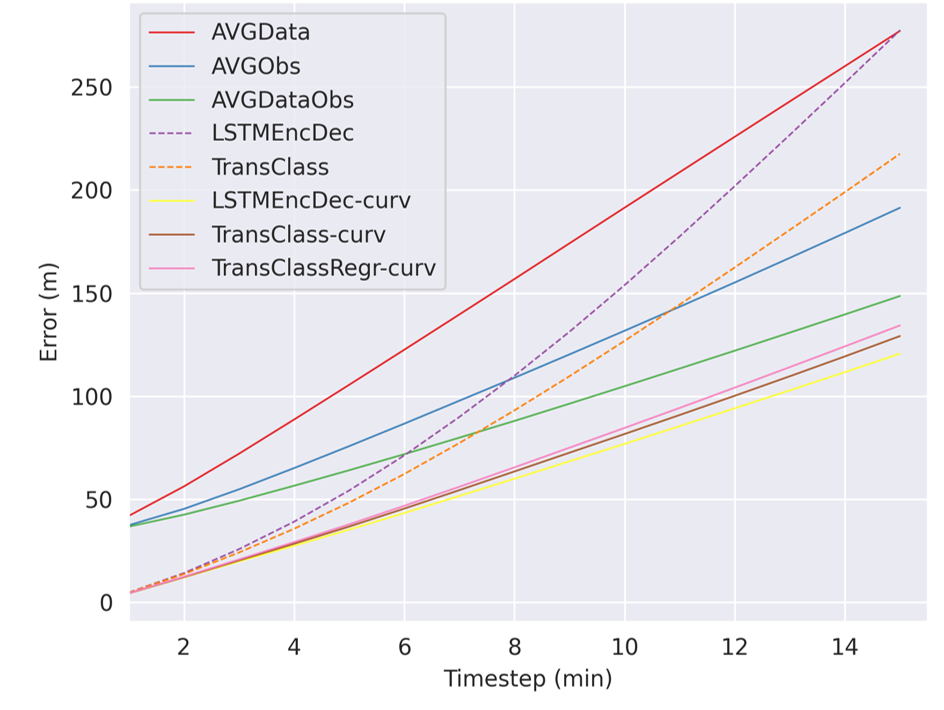}}
	\caption{Mean euclidean distance from target position.}
	\label{fig:result_allModels}
\end{figure}
\begin{table}[]
\centering
\caption{
Error (in meters) after 15 minutes.}\label{tab:errors}
\begin{tabular}{|c|c|c|c|}
\hline
\textbf{Model}      & \textbf{Mean}   & \textbf{$\sigma$}    & \textbf{Median} \\ \hline \hline
AVGDataObs          & 148.57          & 175.51          & 95.60           \\ \hline
LSTMEncDec-curv     & \textbf{120.61} & \textbf{171.09} & \textbf{74.10}  \\ \hline
TransClass-curv     & 129.14          & 173.89          & 83.12           \\ \hline
\end{tabular}
\end{table} 
The LSTMEncDec-curv model reaches a lower validation error and learns faster in terms of required number of epochs as shown in Fig. \ref{fig:ValidTrans}. 
The duration of one training epoch parallelized on two graphical cards are 7-17 min for the LSTMEncDec (depending on $n$ and $m$) and around 2 min for the  transformer model.

\subsection{Comparison to average velocity models}
Three average velocity models are used as baselines. 
The next positions are calculated by iteratively adding a specific distance $d_i$ to the hectometer of the current position $pos_i$ ($i \in {n+1,...,n+m}$) and returning the coordinates of the resulting point on the river axis.
These baseline thus just give a rough estimate of where the vessel is located in terms of how far it gets during the prediction horizon. Note, that for a more sophisticated baseline, the offset of the river axis should be considered as well. The three baseline are:
\begin{enumerate}
    \item Average observed velocity (AVGObs) - $d_i$ is the mean distance travelled during observation and thus constant.
    \item Average hectometer-specific velocity (AVGData) - $d_i$ is the average distance travelled per minute at the current hectometer $hm(pos_i)$, obtained from all vessels passing this hectometer in the training data set. 
    \item Average hectometer-specific velocity with current deviation (AVGDataObs) - $d_i$ is the average hectometer-specific velocity plus the mean deviation of the hectometer-specific velocities during observation. 
\end{enumerate}

\subsection{Prediction results}
The prediction error of the transformer model, its variants, the LSTM encoder-decoder and the baselines is calculated as mean euclidean distance between the predicted and target position per time step and depicted in Fig. \ref{fig:result_allModels} for the case of $n=10$ and $m=15$. Note that the target positions are $pos'_t$ $\forall t \in \{n+2,...,n+m+1\}$ and not the (interpolated) AIS data positions $pos_t$ (see Sec. \ref{sec:approach}). 
The lowest errors are obtained by the context-sensitive models. TransClass-curv is outperformed by LSTMEncDec-curv. After 15 min, the error of LSTMEncDec-curv is 120.61 m (Table \ref{tab:errors}). The addition of a regression sub-module (TransClassRegr-curv) does not improve the prediction. 
The errors of the proposed models and the baselines grow with a nearly constant factor when increasing the prediction horizon, whereas the context-agnostic transformer and LSTM encoder-decoders (TransClass and LSTMEncDec) show an exponential error development and perform worse than the baselines for a prediction horizon of more than 7 and 6 min, respectively. The improvement obtained by the inclusion of context information is higher for LSTMEncDec-curv.

\begin{figure}
\centering
{\includegraphics[height=0.24\textheight,clip,width=0.45\textwidth]{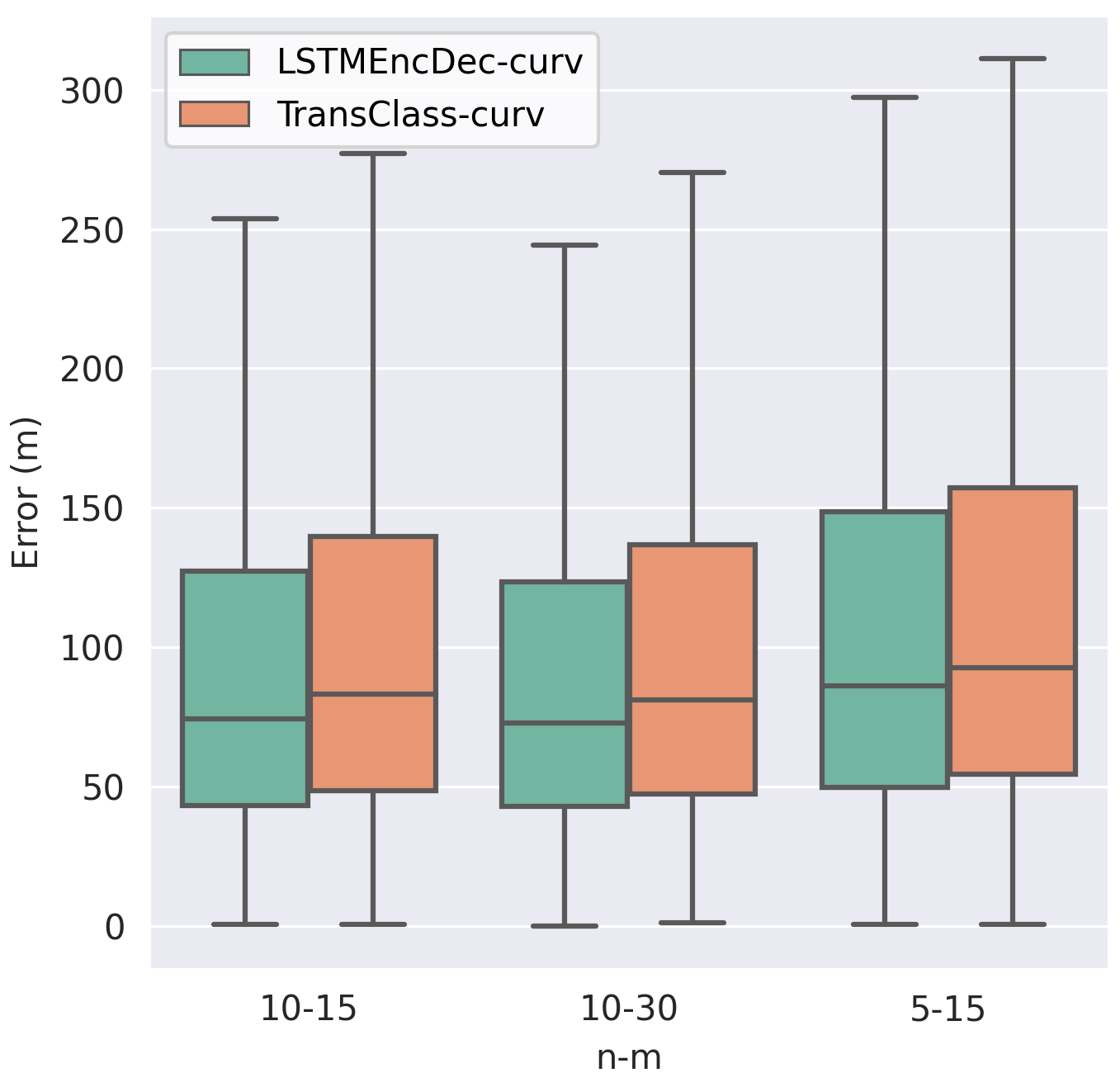}}
	\caption{Distribution of euclidean distances between predicted and target position after 15 min of models trained with varying observation ($n$) and prediction sequence lengths ($m$). 
	}\label{fig:horizon}
\end{figure}

A comparison of the context-sensitive classification models' performance with respect to varying observation and prediction horizons ($n$ and $m$) is given in Fig. \ref{fig:horizon}.  
Decreasing the observation horizon results in higher prediction errors. Training with longer prediction horizons slightly improves the models' performance for the given prediction horizon (15 min). LSTMEncDec-curv outperforms TransClass-curv for all settings. 

An example of a trajectory obtained by LSTMEncDec-curv is visualized in Fig. \ref{fig:predictionEx}. The deviation of the blue trajectory (target data trajectory $pos_1, pos'_2, pos'_3,...,pos'_{n+m+1}$) from the green one (underlying trajectory from AIS obtained after interpolation $pos_1, pos_2,...,pos_{n+m+1}$) is due to the discretization of the distances and COG changes as explained in Sec. \ref{sec:approach}. 

Comparison of research results in vessel trajectory prediction is still challenging due to missing benchmarks and the usage of custom datasets, input/output sequence lengths, temporal resolution, and error measurements throughout the literature. 
Compared to the the multi-task LSTM encoder-decoder proposed in \cite{DijtPimandMettesPascal.2020}, LSTMEncDec-curv and TransClass-curv obtain a  lower ATE\footnote{Absolute trajectory errors, given by the root of the average of the squared euclidean distances between prediction and target position at every time step (error measure used in \cite{DijtPimandMettesPascal.2020}).\label{ate}} after 2.5 min in spite of the lower temporal resolution (see Table \ref{tab:comparison} for details).  
One reason might be the higher input/output sequence length ratio, another that the models herein are optimized for of a specific river section and navigation direction. 

\begin{table}[]
\centering
\caption{Comparison to multi-task encoder decoder\cite{DijtPimandMettesPascal.2020}.}\label{tab:comparison}
\renewcommand{\arraystretch}{1.3}
\begin{adjustbox}{width=0.49\textwidth}
\begin{tabular}{m{2cm}|m{3.7cm}|m{2cm}m{2cm}|}
\cline{2-4}
                                                                     & \textbf{\begin{tabular}[c]{@{}c@{}}Multi-task encoder-decoder\end{tabular}}   & \multicolumn{2}{c|}{\textbf{\begin{tabular}[c]{@{}m{4.5cm}@{}}Context-sensitive encoder-decoders \end{tabular}}}                                            \\ \hline
\multicolumn{1}{|m{2cm}|}{\textbf{Model type}}                                                                   & \begin{tabular}[c]{@{}m{3.7cm}@{}}LSTM encoder-decoder + \linebreak convolutional module\end{tabular} & \multicolumn{1}{m{1.9cm}|}{\begin{tabular}[c]{@{}m{1.9cm}@{}}LSTM\linebreak encoder-decoder\end{tabular}} &  \begin{tabular}[c]{@{}m{1.9cm}@{}}Transformer\linebreak encoder-decoder \end{tabular} 
\\ \hline
\multicolumn{1}{|m{2cm}|}{\textbf{Covered region}}                                                             & \begin{tabular}[c]{@{}m{3.7cm}@{}}Germany, Belgium, Netherlands\end{tabular}               & \multicolumn{2}{c|}{\begin{tabular}[c]{@{}m{4.5cm}@{}}Danube km 2245-2321.2, uphill\end{tabular}}                                                       \\ \hline
\multicolumn{1}{|m{2cm}|}{\textbf{Data source}}                                                                  & AIS, radar images, ENC                                                                      & \multicolumn{2}{m{4.5cm}|}{AIS}                                                                                                                                     \\ \hline
 \multicolumn{1}{|m{2cm}|}{\textbf{Dataset size}}                                                               & 44 h trajectories                                                                      & 
 \multicolumn{2}{c|}{\begin{tabular}[l]{@{}m{4.5cm}@{}}272,573 sub-trajectories  \hspace{1.5cm} \linebreak      ($n=10$, $m=15$)\end{tabular}}                                                       \\ \hline

\multicolumn{1}{|m{2cm}|}{\textbf{Time step size}}                                                             & 3 s                                                                                  & \multicolumn{2}{m{4.5cm}|}{1 min}                                                                                                                                   \\ \hline
\multicolumn{1}{|m{2cm}|}{\textbf{\begin{tabular}[c]{@{}m{2cm}@{}}Input/output\linebreak sequence length\end{tabular}}} & \begin{tabular}[c]{@{}m{2.7cm}@{}}5/50 (15 s/2.5 min)\end{tabular}                       & \multicolumn{2}{c|}{\begin{tabular}[c]{@{}m{4.5cm}@{}}10/15 (10 min/15 min)\end{tabular}}                                                                      \\ \hline
\multicolumn{1}{|m{2cm}|}{\textbf{ATE\footref{ate} after 150 s}}                                                            & 17.13 m                                                                                 & \multicolumn{1}{m{1.9cm}|}{\begin{tabular}[c]{@{}m{1.9cm}@{}}\textbf{11.71 m}\end{tabular}} & \begin{tabular}[c]{@{}m{1.9cm}@{}}\textbf{11.88 m} \end{tabular} \\ \hline
\end{tabular}
\end{adjustbox}
\end{table}
	
\section{CONCLUSION}\label{sec:conclusion}
Different encoder-decoder models are used for predicting the time series of the next 15 min of a vessel's trajectory after having observed 5 to 10 min.  
The LSTM encoder-decoder outperforms the transformer model, but is computationally more expensive. Using a context vector as first decoder input that represents the radii of the river section the vessel is heading the performance is significantly improved. 
The classification reformulation of the regression problem is necessary to successfully train the transformer model. The error introduced by the discretization of distance and COG change is assumed to be neglectable for the prediction horizon under consideration, but would be unacceptable in applications with high precision requirements, e.g. collision avoidance maneuver prediction. It should, however, be ensured that the positions obtained after discretization are located in the navigable area and not outside, which is theoretically what is possible with the current implementation. 
The proposed models outperform the baselines and the multi-task encoder-decoder \cite{DijtPimandMettesPascal.2020}. 
In future works, the effect of enriching the context vector with further waterway-specific information needs to be investigated. Moreover, substituting the greedy decoding method used to generate the output sequence from the output probabilities by a beam search decoding could possibly lead to better results. Sampling output sequences would be required to get an estimate of the model's uncertainty. Finally, due to the sensitivity of deep neural networks to hyper-parameters, further hyper-parameter optimization needs to be conducted. 

\addtolength{\textheight}{-12cm}   
\bibliographystyle{IEEEtran}
\bibliography{IEEEexample}

\begin{thebibliography}{10}
\providecommand{\url}[1]{#1}
\csname url@samestyle\endcsname
\providecommand{\newblock}{\relax}
\providecommand{\bibinfo}[2]{#2}
\providecommand{\BIBentrySTDinterwordspacing}{\spaceskip=0pt\relax}
\providecommand{\BIBentryALTinterwordstretchfactor}{4}
\providecommand{\BIBentryALTinterwordspacing}{\spaceskip=\fontdimen2\font plus
\BIBentryALTinterwordstretchfactor\fontdimen3\font minus
  \fontdimen4\font\relax}
\providecommand{\BIBforeignlanguage}[2]{{%
\expandafter\ifx\csname l@#1\endcsname\relax
\typeout{** WARNING: IEEEtran.bst: No hyphenation pattern has been}%
\typeout{** loaded for the language `#1'. Using the pattern for}%
\typeout{** the default language instead.}%
\else
\language=\csname l@#1\endcsname
\fi
#2}}
\providecommand{\BIBdecl}{\relax}
\BIBdecl

\bibitem{Murray.2021b}
B.~Murray and L.~P. Perera, ``{An {AIS}-based deep learning framework for
  regional ship behavior prediction},'' \emph{Reliability Engineering {\&}
  System Safety}, vol. 124, no.~5, p. 107819, 2021.

\bibitem{Capobianco.2021}
S.~Capobianco, L.~M. Millefiori, N.~Forti, P.~Braca, and P.~Willett, ``{Deep
  Learning Methods for Vessel Trajectory Prediction based on Recurrent Neural
  Networks},'' \emph{IEEE Transactions on Aerospace and Electronic Systems},
  vol.~57, no.~6, pp. 4329--4346, 2021.

\bibitem{You.2020}
L.~You, S.~Xiao, Q.~Peng, C.~Claramunt, X.~Han, Z.~Guan, and J.~Zhang,
  ``{ST-Seq2Seq: A Spatio-Temporal Feature-Optimized Seq2Seq Model for
  Short-Term Vessel Trajectory Prediction},'' \emph{IEEE Access}, vol.~8, pp.
  218\,565--218\,574, 2020.

\bibitem{Nguyen.2021TrAISformerAGT}
D.~Nguyen and R.~Fablet, ``Tr{AIS}former-a generative transformer for ais
  trajectory prediction,'' \emph{ArXiv}, vol. abs/2109.03958, 2021.

\bibitem{Rudenko.2020}
A.~Rudenko, L.~Palmieri, M.~Herman, K.~M. Kitani, D.~M. Gavrila, and K.~O.
  Arras, ``{Human motion trajectory prediction: A survey},'' \emph{The
  International Journal of Robotics Research}, vol.~39, no.~8, pp. 895--935,
  2020.

\bibitem{Mozaffari.2020}
S.~Mozaffari, O.~Y. Al-Jarrah, M.~Dianati, P.~Jennings, and A.~Mouzakitis,
  ``{Deep Learning-Based Vehicle Behavior Prediction for Autonomous Driving
  Applications: A Review},'' \emph{IEEE Transactions on Intelligent
  Transportation Systems}, vol.~23, no.~1, pp. 33--47, 2022.

\bibitem{Tu.2018}
E.~Tu, G.~Zhang, L.~Rachmawati, E.~Rajabally, and G.-B. Huang, ``{Exploiting
  AIS Data for Intelligent Maritime Navigation: A Comprehensive Survey From
  Data to Methodology},'' \emph{IEEE Transactions on Intelligent Transportation
  Systems}, vol.~19, no.~5, pp. 1559--1582, 2018.

\bibitem{Azimi.2020}
S.~Azimi, J.~Salokannel, S.~Lafond, J.~Lilius, M.~Salokorpi, and I.~Porres,
  ``{A survey of machine learning approaches for surface maritime
  navigation},'' in \emph{{Maritime Transport VIII: Proceedings of the 8th
  International Conference on Maritime Transport (Maritime Transport '20)}},
  Barcelona, 2020, pp. 103--117.

\bibitem{Xiao.2020}
Z.~Xiao, X.~Fu, L.~Zhang, and R.~S.~M. Goh, ``{Traffic Pattern Mining and
  Forecasting Technologies in Maritime Traffic Service Networks: A
  Comprehensive Survey},'' \emph{IEEE Transactions on Intelligent
  Transportation Systems}, vol.~21, no.~5, pp. 1796--1825, 2020.

\bibitem{Jin.2021}
J.~Jin, W.~Zhou, and B.~Jiang, ``{An Overview: Maritime Spatial-Temporal
  Trajectory Mining},'' \emph{Journal of Physics: Conference Series}, vol.
  1757, no.~1, p. 012125, 2021.

\bibitem{Nguyen.2018}
D.-D. Nguyen, C.~{Le Van}, and M.~I. Ali, ``{Vessel Trajectory Prediction using
  Sequence-to-Sequence Models over Spatial Grid},'' in \emph{{Proceedings of
  the 12th ACM International Conference on Distributed and Event-based Systems
  (DEBS '18)}}, New York, 2018, pp. 258--261.

\bibitem{Forti.2020}
N.~Forti, L.~M. Millefiori, P.~Braca, and P.~Willett, ``{Prediction of Vessel
  Trajectories From AIS Data Via Sequence-To-Sequence Recurrent Neural
  Networks},'' in \emph{{2020 IEEE Int. Conference on Acoustics, Speech and
  Signal Processing}}, Barcelona, 2020, pp. 8936--8940.

\bibitem{Sekhon.2020}
J.~Sekhon and C.~Fleming, ``{A Spatially and Temporally Attentive Joint
  Trajectory Prediction Framework for Modeling Vessel Intent},'' in
  \emph{{Proceedings of the 2nd Conference on Learning for Dynamics and
  Control, PMLR}}, 2020, vol. 120, pp. 318--327.

\bibitem{DijtPimandMettesPascal.2020}
P.~Dijt and P.~Mettes, ``{Trajectory Prediction Network for Future Anticipation
  of Ships},'' in \emph{{Proceedings of the 2020 International Converence on
  Multimedia Retrieval}}, Dublin, 2020, pp. 73--81.

\bibitem{Yuan.2020}
Z.~Yuan, J.~Liu, Y.~Liu, Q.~Zhang, and R.~W. Liu, ``{A multi-task analysis and
  modelling paradigm using LSTM for multi-source monitoring data of inland
  vessels},'' \emph{Ocean Engineering}, vol. 213, p. 107604, 2020.

\bibitem{Vaswani.2017}
A.~Vaswani, N.~Shazeer, N.~Parmar, J.~Uszkoreit, L.~Jones, A.~N. Gomez,
  {\L}.~Kaiser, and I.~Polosukhin, ``{Attention is all you need},'' in
  \emph{{Proceedings of the 31st International Conference on Neural Information
  Processing Systems}}, vol.~30, Long Beach, USA, 2017, pp. 6000--6010.

\bibitem{Cai.2020}
L.~Cai, K.~Janowicz, G.~Mai, B.~Yan, and R.~Zhu, ``{Traffic transformer:
  Capturing the continuity and periodicity of time series for traffic
  forecasting},'' \emph{Transactions in GIS}, vol.~24, no.~3, pp. 736--755,
  2020.

\bibitem{Chen.2021}
W.~Chen, F.~Wang, and H.~Sun, ``{S2TNet: Spatio-Temporal Transformer Networks
  for Trajectory Prediction in Autonomous Driving},'' in \emph{Proceedings of
  the 13th Asian Conference on Machine Learning}, vol. 157, 2021, pp. 454--469.

\bibitem{Yu.2020}
C.~Yu, X.~Ma, J.~Ren, H.~Zhao, and S.~Yi, ``{Spatio-Temporal Graph Transformer
  Networks for Pedestrian Trajectory Prediction},'' in \emph{{Computer Vision
  – ECCV 2020, 16th European Conference, Proceedings, Part {XII}}}, Glasgow,
  UK, 2020, pp. 507--523.

\bibitem{Sutskever.2014}
I.~Sutskever, O.~Vinyals, and Q.~V. Le, ``{Sequence to Sequence Learning with
  Neural Networks},'' in \emph{{Proceedings of the 27th Int. Conference on
  Neural Information Processing Systems}}, vol.~2, Cambridge, USA, 2014, p.
  3104–3112.

\end{thebibliography}

\end{document}